\documentclass[10pt, a4paper, oneside]{article}
\usepackage[final]{lrec-coling2024} 
\usepackage{makecell}
\usepackage{subfiles}
\usepackage{natbib}
\usepackage{multibib}
\usepackage[nottoc]{tocbibind}
\usepackage{multirow}
\makeatletter
\def\@mb@citenamelist{cite,citep,citet,citealp,citealt,citepalias,citetalias}
\makeatother
\newcites{languageresource}{~}

\usepackage{graphicx}
\usepackage{tabularx}
\usepackage{soul}
\usepackage{booktabs}

\usepackage{hyphenat}
\usepackage{array}
\usepackage[normalem]{ulem}
\useunder{\uline}{\ul}{}

\usepackage{bm}

\newcommand{\CorpusAbbrev}{CAMIR}

\usepackage{titlesec}

\usepackage{xcolor}
\usepackage{hyperref}
 \definecolor{darkblue}{rgb}{0, 0, 0.5}
  \hypersetup{colorlinks=true, citecolor=darkblue, linkcolor=darkblue, urlcolor=darkblue}

\usepackage{xstring}

\usepackage{color}

\title{\textbf{A Novel Corpus of Annotated Medical Imaging Reports and Information Extraction Results Using BERT-based Language Models}}

\name{ \parbox{\textwidth}{\fontsize{12}{0}\selectfont\centering
Namu Park\textsuperscript{1}\textsuperscript{*},
Kevin Lybarger\textsuperscript{2}\textsuperscript{*},
Giridhar Kaushik Ramachandran\textsuperscript{2}, \\
Spencer Lewis\textsuperscript{3}, 
Aashka Damani\textsuperscript{4},
Özlem Uzuner\textsuperscript{2},
Martin Gunn\textsuperscript{5}, \\
Meliha Yetisgen\textsuperscript{1}
}}

\address{\textsuperscript{1}Department of Biomedical Informatics \& Medical Education, University of Washington\\
\textsuperscript{2}Department of Information Sciences and Technology, George Mason University\\
\textsuperscript{3}Department of Radiology, School of Medicine, Stanford University\\
\textsuperscript{4}School of Medicine, University of Washington\\
\textsuperscript{5}Department of Radiology, School of Medicine, University of Washington\\
         \textsuperscript{1,4,5}Seattle, WA, USA, 
         \textsuperscript{2}Fairfax, VA, USA,
         \textsuperscript{3}Stanford, CA, USA
         \\
         \{npark95, aashkad, marting, melihay\}@uw.edu\\
         \{klybarge, gramacha, ouzuner\}@gmu.edu\\ 
         lewispen@stanford.edu\\
         \textsuperscript{*}Authors contributed equally to this paper.
         }

\abstract{Medical imaging is critical to the diagnosis, surveillance, and treatment of many health conditions, including oncological, neurological, cardiovascular, and musculoskeletal disorders, among others. Radiologists interpret these complex, unstructured images and articulate their assessments through narrative reports that remain largely unstructured. This unstructured narrative must be converted into a structured semantic representation to facilitate secondary applications such as retrospective analyses or clinical decision support. Here, we introduce the Corpus of Annotated Medical Imaging Reports (CAMIR), which includes 609 annotated radiology reports from three imaging modality types: Computed Tomography, Magnetic Resonance Imaging, and Positron Emission Tomography-Computed Tomography. Reports were annotated using an event-based schema that captures clinical indications, lesions, and medical problems. Each event consists of a trigger and multiple arguments, and a majority of the argument types, including anatomy, normalize the spans to pre-defined concepts to facilitate secondary use. CAMIR uniquely combines a granular event structure and concept normalization. To extract CAMIR events, we explored two BERT (Bi-directional Encoder Representation from Transformers)-based architectures, including an existing architecture (mSpERT) that jointly extracts all event information and a multi-step approach (PL-Marker++) that we augmented for the CAMIR schema.
\\ \newline}

\begin{document}

\maketitleabstract

\section{Introduction}


Radiology reports document radiologists' interpretation of medical images through detailed narrative text. Although some studies have explored structured reports that utilize common data elements to express radiologists' interpretations through pre-defined medical concepts \cite{rubin2017common}, the majority of radiology reports utilize narrative text \cite{willemink2020preparing}. Information extraction (IE) techniques can automatically convert unstructured reports to structured semantic representations to allow utilization of the textual information in secondary-use applications. Example applications include cohort discovery \cite{casey_systematic_2021}, epidemiology \cite{casey_systematic_2021}, image retrieval \cite{gerstmair2012intelligent}, automated follow-up tracking \cite{mabotuwana2019automated}, computer-vision applications \cite{zech2018natural}, decision support \cite{demner2009can}, and report summarization \cite{wiggins2021natural}.

Although there is a well-established body of radiology IE research, most of this research focuses on specific clinical tasks \cite{casey_systematic_2021, Donnelly2022} or medical conditions, utilizes a single imaging modality, or implements an annotation schema that does not comprehensively capture the available information. 
To address these limitations, we introduce a novel annotated corpus, the \textit{Corpus of Annotated Medical Imaging Reports} (CAMIR), that is relevant to a broad set of applications. CAMIR includes Computed Tomography (CT), Magnetic Resonance Imaging (MRI), and Positron Emission Tomography-Computed Tomography (PET-CT) reports. The reports are annotated using a granular event schema, where clinical indication, lesion, and medical problem findings are characterized through multiple arguments, including assertion values (present vs. absent), normalized anatomy using a hierarchical ontology of 87 SNOMED-CT concepts, and other clinically important attributes. CAMIR includes 609 annotated radiology reports with 1,494 indication events, 5,709 lesion events, and 6,255 medical problem events. CAMIR has a high inter-annotator agreement (>0.805 F1) for event triggers and an overall inter-annotator agreement of 0.762 F1. We present initial IE results using two BERT (Bi-directional Encoder Representation from Transformers)-based language models trained and evaluated on CAMIR, including the Multi-label Span-based Entity and Relation Transformer (mSpERT) \cite{eberts2020span,lybarger2022leveraging} and an augmented version of Packed-levitated Markers \cite{ye2022packed} (referred to as PL-Marker++). Both architectures achieve performance comparable to the inter-annotator agreement (IAA), with PL-Marker++ achieving the highest overall performance.

\section{Related Work}

There is a significant body of research that explores IE within the radiology domain, including the creation of annotated corpora and the development of extraction models \cite{pons2016natural, casey_systematic_2021, lopez2022natural}. In this section, we discuss existing research in clinical NLP focusing on radiology corpora and relevant IE techniques.

\subsection{Radiology Corpora}

Radiology reports present nuanced and complex descriptions of medical findings, which existing annotated corpora capture with varying degrees of granularity.
Document-level or sentence-level annotations map relevant text to normalized values, targeting diverse label categories such as metastases characteristics \cite{do_patterns_2021} and incidental findings \cite{trivedi_interactive_2019}.  Entity annotations identify phrases of interest, capturing concepts like anatomical location \cite{wang2019natural} or tumor attributes \cite{yim2016tumor}. Relation and event annotations enable more nuanced representations, like the multi-attribute characterization of medical problems \cite{lau2022event}. Selected studies have integrated the normalization of radiological concepts related to anatomy \cite{lybarger_extracting_2021, datta2022fine, nishigaki2023bert} and other radiology terminology \cite{datta2020radlex}.

Existing corpora often exhibit limitations in various dimensions, such as the diversity of the patient populations represented, the range of imaging modalities included, the scale of the datasets, or the granularity and comprehensiveness of the annotation schemas employed. Some studies concentrate on specific diseases or conditions like hepatocellular carcinoma \cite{yim2016tumor} or appendicitis \cite{rink2013extracting}, limiting the represented patient populations. Others are limited to single imaging modalities \cite{lau2022event, sugimoto_extracting_2021} or small corpora (n<200) \cite{hassanpour_information_2016}. 
Other relevant relation extraction work does not include the normalization of extracted spans to key concepts \cite{jain_radgraph_2021}. More recent work \cite{lybarger_extracting_2021}  extracts findings with the associated normalized anatomy values; however, the findings are not fully characterized through granular attributes. 

To our knowledge, CAMIR is the first annotated corpus to uniquely combine clinical concept normalization with granular event annotations to comprehensively capture important clinical findings. Additionally, CAMIR includes a diverse set of reports from three imaging modalities that were sampled from all patients at the University of Washington (UW). CAMIR's fine-grained annotation schema with concept normalization and heterogeneous set of reports can support a wide range of secondary-use applications.

\subsection{IE Methods in Radiology}

Early radiology IE research employed discrete machine learning models with engineered features. For instance, Support Vector Machines were used to detect appendicitis findings \cite{rink2013extracting} and Conditional Random Fields were utilized to extract anatomy and findings \cite{hassanpour_information_2016}. These discrete modeling approaches were supplanted by neural network architectures, such as Convolutional Neural Network and Recurrent Neural Networks. These neural architectures outperform their predecessors in many radiology IE tasks, including but not limited to recommendation extraction \cite{carrodeguas2019use, steinkamp2021automatic}, clinical concept identification \cite{zhu2018clinical}, and spatial information extraction \cite{datta2020understanding}. Currently, pre-trained Language Models dominate the IE landscape in radiology, similar to other domains. BERT \cite{devlin2019bert} models have been extensively implemented for tasks ranging from observation detection \cite{irvin_chexpert_2019} to anatomy classification \cite{nishigaki2023bert} and relation-based finding extraction \cite{lybarger_extracting_2021}. Most recently, Generative Pre-trained Transformers (GPT) models are being leveraged to extract structured information from radiology reports \cite{fink2023potential, mukherjee2023feasibility, adams2023leveraging}. In this paper, we present the extraction results of two high-performing BERT-based models, which were tailored to reflect the granularity of our annotation schema and serve as a foundation upon which future work can build.

\section{Methods}

\begin{figure*}[!ht]
  \begin{center}
    \includegraphics[width=6.2in]{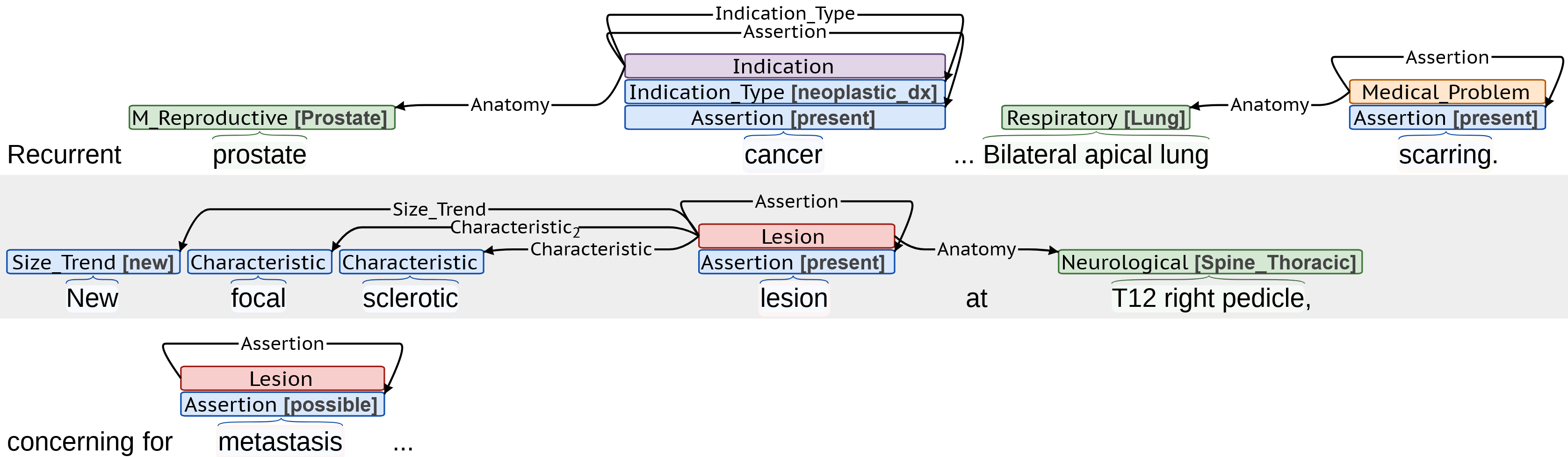}
  \end{center}
  \caption{Examples of sentences annotated with event schema}
  \label{annotation_example}
\end{figure*}

\subsection{Corpus Creation}

We used an existing clinical database of 1,417,586 CT, 541,388 MRI, and 39,150 PET-CT reports from 2007-2020 which includes the general patient population from four UW Medical System hospitals. We randomly sampled reports from each modality: 203 CT, 202 MRI, and 204 PET-CT. The reports were automatically de-identified using a neural de-identifier \cite{leeDobbins2021}. The study was approved by the UW Institutional Review Board (IRB).

\subsubsection{Annotation Schema}

\begin{table*}[htb!]
    \small
    \centering
    \renewcommand{\arraystretch}{0.9}

\begin{tabular}{m{1.2cm} m{1.8cm} m{7.6cm} m{3.5cm}}
\toprule
\textbf{Event}                          & \textbf{Trigger/ \mbox{Argument}} & \textbf{Argument subtypes}                                                     & \textbf{Span examples}                                \\ \toprule
\multirow{4}{*}{Indication}             & Trigger\textsuperscript{*}        & --                                                                                    & ``hemorrhage," ``sepsis"                              \\ \cmidrule{2-4}
& Type\textsuperscript{*}& \{trauma, symptom, neoplastic diagnosis, non-neoplastic diagnosis\}     & ``seminoma," ``sarcoid"                               \\ \cmidrule{2-4}
                                        & Assertion\textsuperscript{*}      & \{present, absent, possible\}                                                         & ``r/o," ``concern"                                    \\ \cmidrule{2-4}
                                        & Anatomy\textsuperscript{}         & \mbox{Anatomy Parent and Child labels}\textsuperscript{+}                     & ``abdominal," ``alveolar"                             \\ \midrule
\multirow{7}{*}{Lesion}                 & Trigger\textsuperscript{*}        & --                                                                                    & ```lymhpadenopathy"                      \\ \cmidrule{2-4}
                                        & Assertion\textsuperscript{*}      & \{present, absent, possible\}                                                         & \mbox{``most likely,"}  \\ \cmidrule{2-4}
                                        & Anatomy                           & \mbox{Anatomy Parent and Child labels}\textsuperscript{+}                     &  ``lower back"        \\ \cmidrule{2-4}
& Size                              &  \{current, past\}                                                                    & \mbox{``up to 5mm"}                \\ \cmidrule{2-4}
                                        & Size Trend                        & \{{new, disappear, increasing, decreasing, \mbox{no-change}}\}                        & \mbox{``increasing in size"}            \\ \cmidrule{2-4}
                                        & Count                             & --                                                                                    & ``multiple," \mbox{``numerous"}                       \\ \cmidrule{2-4}
                                        & Characteristic                    & --                                                                                    & ``peripheral," ``enlarged"                            \\ \midrule
\multirow{3}{1.2cm}{Medical Problem}    & Trigger\textsuperscript{*}        & --                                                                                    & ``dilation," ``calcification"                         \\ \cmidrule{2-4}
                                        & Assertion\textsuperscript{*}      & \{\nohyphens{present, absent, possible}\}                                             &  ``possibly"      \\ \cmidrule{2-4}
                                        & Anatomy                           & \mbox{Anatomy Parent and Child labels}\textsuperscript{+}                     & ``mucosal," ``supraagger"                             \\ \bottomrule                                                        
\end{tabular}
    
    \caption{Summary of the event schema. \textsuperscript{*} indicates the argument is required. \textsuperscript{+} \textit{Anatomy Parent} and \textit{Anatomy Child} are list in Table \ref{anatomy_hierarchy}. ``Dx" refers to diagnosis.}
    \label{annotated_phenomena}
\end{table*}

In CAMIR’s event schema, each event includes a trigger that identifies the event and arguments that characterize the event. Table \ref{annotated_phenomena} summarizes the schema, and Figure \ref{annotation_example} presents annotation examples from the BRAT rapid annotation tool \cite{stenetorp2012brat}, which was used throughout the annotation process. CAMIR includes three event types: (1) \textit{Indication} - reason for the imaging (e.g., “cancer” in line 1 of Figure \ref{annotation_example}), (2) \textit{Lesion} – mass-occupying pathological structures (e.g., “metastasis” in line 3 of Figure \ref{annotation_example}); and (3) \textit{Medical Problem} - non-mass-like abnormalities, defined as a finding that is not a potential mass (e.g., “scarring” in line 1 of Figure \ref{annotation_example}). There are two argument types: (1) \textit{span-only} arguments assign text spans an argument label (e.g., “focal” assigned \textit{Characteristic} argument in line 2 of Figure \ref{annotation_example}) and (2) \textit{span-with-value} arguments assign text spans both an argument label and argument subtype label (e.g., “New” assigned \textit{Size Trend} argument with subtype value \textit{new} in line 2 of Figure \ref{annotation_example}). To improve the granularity of our annotation schema, anatomy arguments are normalized to a set of hierarchical anatomical SNOMED-CT concepts, including 16 \textit{Anatomy Parent} and 71 \textit{Anatomy Child} labels listed in Table \ref{anatomy_hierarchy} (e.g., “Bilateral apical lung” assigned \textit{Anatomy Parent} - \textit{Respiratory} and \textit{Anatomy Child} - \textit{Lung} in line 1 of Figure \ref{annotation_example}). 

 \begin{table*}
     \small
     \centering
     \renewcommand{\arraystretch}{0.9}
\begin{tabular}{m{3.0cm} m{12.0cm}}
\toprule
\textbf{Anatomy Parent} &   \textbf{Anatomy Children} \\ \midrule
Abdomen &  Abdominal Wall, Adrenal Gland, Mesentery, Peritoneal Sac, Retroperitoneal, \& Spleen   \\ \midrule
Body Regions &   Entire Body, Lower Limb, Pelvis, \& Upper Limb  \\ \midrule
Cardiovascular &   Arterial, Coronary Artery, Heart, Pericardial Sac, Pulmonary Artery, \& Venous  \\ \midrule
Digestive &   Esophagus, Intestine, Large Intestine, Small Intestine, \& Stomach  \\ \midrule
Female Reproductive \& Obstetric &  Adnexal, Breast, Extra-embryonic, Female Genital Structure, Fetus, Ovary, Placenta, Umbilical Cord, \& Uterus  \\ \midrule
Head \& Neck &   Ear, Eye, Laryngeal, Mouth, Nasal Sinus, Neck, Pharynx, \& Thyroid  \\ \midrule
Hepato-Biliary & Bile Duct, Gallblader, Liver, \& Pancreas  \\ \midrule
Lymphatic &   --  \\ \midrule
Male Reproductive &   Epididymis, Prostate, \& Testis  \\ \midrule

Musculo-Skeletal &   Bone/Joint, \& Skeletal and/or Smooth Muscle  \\ \midrule
Neurological &   Brain, Cerebrospinal Fluid Pathway, Cerebrovascular System, Extraaxial, Nerve, Pituitary, \& Spine -  Cervical, Cord, Lumbar, Sacral, Thoracic, or Unspecified  \\ \midrule
Respiratory &   Lung, Pleural Membrane, \& Tracheobronchial  \\ \midrule
Skin &   Skin and or Mucous Membrane, \& Subcutaneous  \\ \midrule 
Thoracic &   Mediastinal  \\ \midrule
Urinary &   Kidney, Ureter, \& Urinary Bladder  \\ 
\midrule
Miscellaneous & Adipose Tissue, Biomedical Device, \& Connective Tissue  \\ 
\bottomrule
\end{tabular}
     \caption{Anatomy Parent-Child Hierarchy. All 16 Parent and 71 Child labels map to SNOMED-CT concepts, but label names are shortened for space. All Parent labels include an \textit{Undetermined} child label.}
     \label{anatomy_hierarchy}
\end{table*}

\subsubsection{Annotation Process}

Four medical students annotated CAMIR. A senior radiology resident and an experienced board-certified attending radiologist provided domain expertise in creating the annotation guidelines and resolving the ambiguities during annotation. The annotation guidelines were designed with the efforts of a medical resident and a board-certified radiologist with 20+ years of experience and profound knowledge of clinical NLP. After a series of meetings, we updated the annotation guidelines multiple times to ensure the guidelines accurately and comprehensively capture the indication, finding, and lesion information relevant to a wide range of clinical research, including our current exploration of cancer and incidental findings. The hierarchical anatomy normalization schema was developed with the help of a board-certified radiologist by reflecting the widely used SNOMED-CT concepts.
Two pairs of two medical students doubly annotated 357 reports, and 252 reports were single-annotated by the same annotators. Annotators reached a consistent level of IAA after 5 rounds of double annotation. Disagreements were adjudicated with the help of domain experts who created and revised the annotation guidelines when needed. We then transitioned to single annotation for the next four rounds to expedite the annotation process. CAMIR includes training, validation, and test set splits (70\%:10\%:20\%). The training set is 41\% doubly annotated, and the entire validation and test sets are doubly annotated to ensure evaluation reliability. 

We singly annotated the training set to create a larger and more diverse training set, while providing the most robust data set for the validation and test sets using double-annotation. The consistency of annotations between singly annotated and doubly annotated reports was evaluated by analyzing the average frequency of labels per report. The doubly annotated reports have an average of 2.65±0.48 Indication, 10.15±1.31 Medical Problem, and 9.77±0.99 Lesion triggers per report, and the singly annotated reports include an average of 2.14±0.26 Indication, 9.91±2.58 Medical Problem, and 8.78±1.06 Lesion triggers per report. 
The frequency of triggers is slightly lower in the singly annotated reports, suggesting there is some reduced annotation recall for the singly annotated reports; however, the evaluation was performed on the doubly annotated test set, and any annotation noise associated with the singly training examples is captured by this evaluation.

\subsection{Extraction Architectures}

To extract the \CorpusAbbrev{} events, we explored two state-of-the-art BERT \cite{devlin2019bert}-based Language Models: (1) mSpERT \cite{eberts2020span, lybarger2022leveraging} and (2) an augmented version of PL-Marker \cite{ye2022packed} referred to as PL-Marker++. For both systems, we decomposed events into a set of entities and relations, where the relation head is a trigger and the relation tail is an argument.

\subsubsection{mSpERT}

\begin{figure*}[ht]
  \centering
    \includegraphics[width=6.0in]{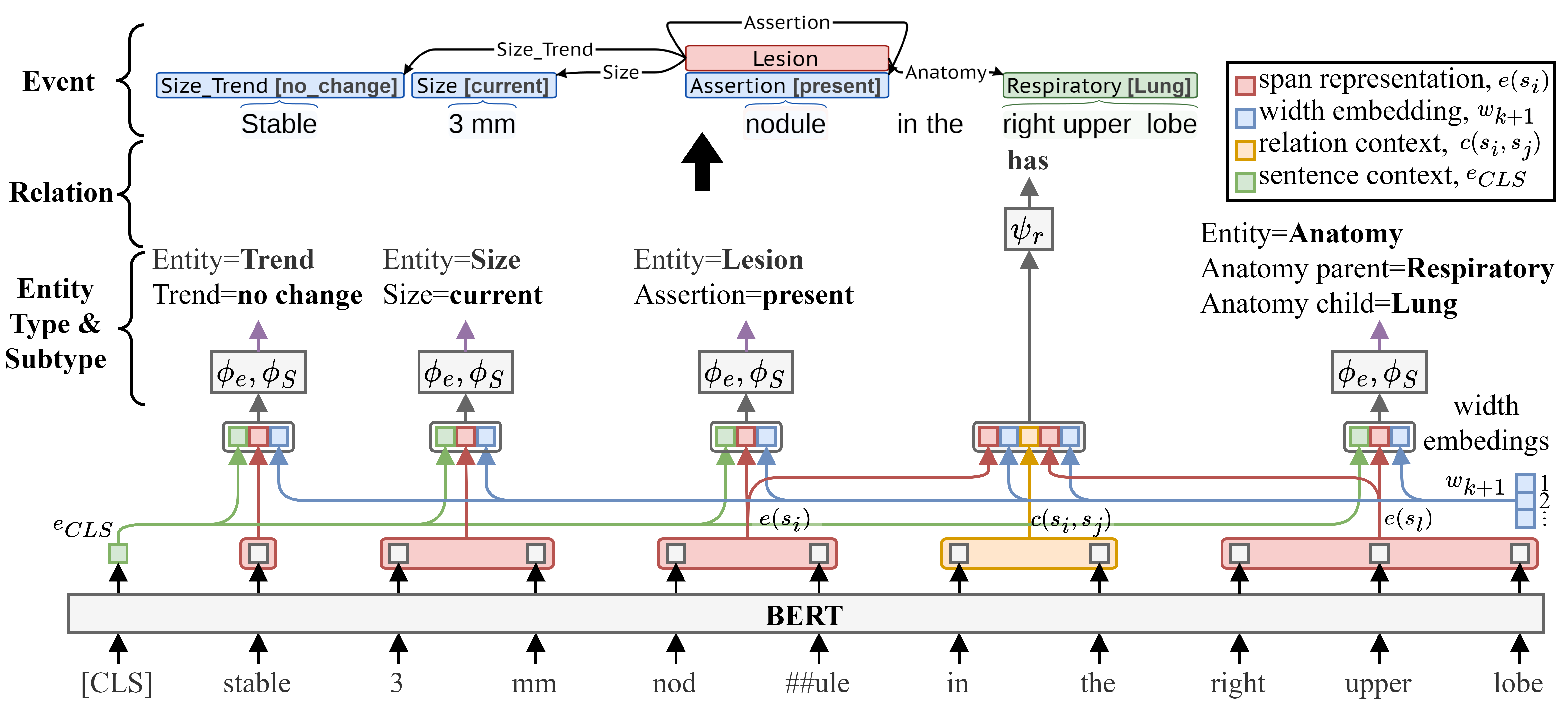}
  \caption{mSpERT framework}
  \label{mspert}
\end{figure*}

SpERT \cite{eberts2020span} jointly extracts entities and relations using BERT \cite{devlin2019bert} with output layers that classify spans and predict relations. mSpERT \cite{lybarger2022leveraging} includes additional output layers to allow multi-label span predictions, which we use to predict subtype labels. Figure \ref{mspert} shows the mSpERT architecture, which includes Entity Type, Entity Subtype, and Relation output layers. The Entity Type and Relation layers of mSpERT are identical to the original SpERT implementation, and the Entity Subtype layer allows multi-label span predictions. The Entity Type classifier ($\phi_e$) is a linear layer that operates on the sentence representation ($e_{CLS}$), max-pooled span hidden states ($e(s_i)$), and learned span width embeddings ($w_{k+1}$). The Entity Subtype classifiers ($\phi_S$) are separate linear layers for each \textit{span-with-value} argument that operate on the same input as the Entity Type classifier but also incorporate the Entity Type logits. The Relation classifier ($\psi_r$) predicts links between entity spans using a linear layer that operates on max-pooled spans ($e(s_i)$ and ($e(s_l)$), span width embeddings ($w_{k+1}$), and max-pooled hidden states between the entity spans ($c(s_i, s_j)$). The mSpERT predictions can generate the \CorpusAbbrev{} event structure. 


\subsubsection{PL-Marker++}

\begin{figure*}[htb!]
  \begin{center}
    \includegraphics[width=6.0in]{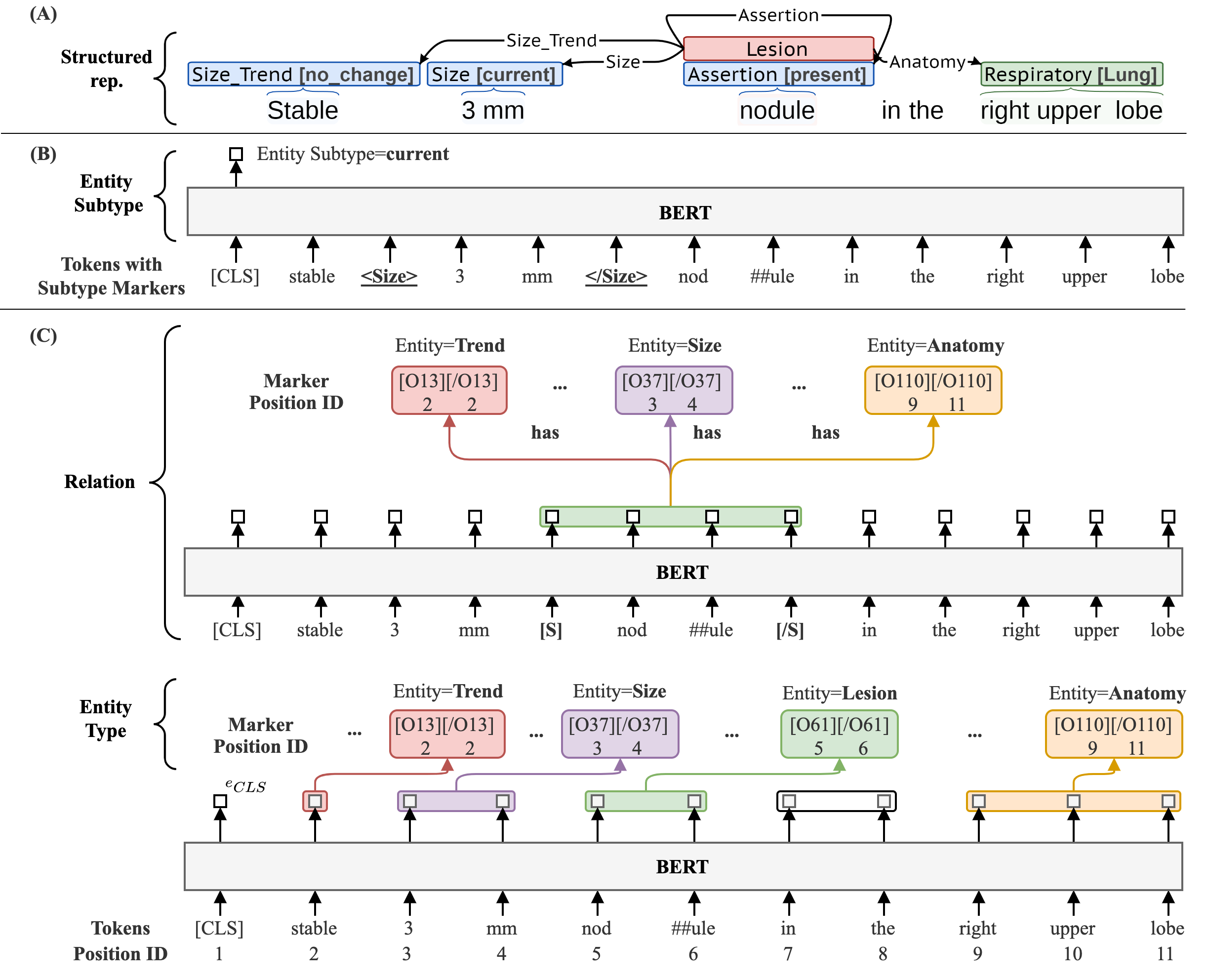}
  \end{center}
  \caption{PL-Marker++ framework}
  \label{plmarker}
\end{figure*}

PL-Marker \cite{ye2022packed} is a multi-stage extraction framework, where the first stage identifies entities and second stage resolves relations. To extract CAMIR events, we introduced an augmented version of PL-Marker, referred to as PL-Marker++, which includes a third classification stage for the \textit{span-with-value} subtype labels. Figure 3 presents the PL-Marker++ architecture, where the Entity Type and Relation stages are identical to the original PL-Marker model. The Entity Type stage uses a group packing approach to process many spans concurrently while considering their interdependencies. The Relation stage uses a subject-oriented packing strategy to pack each relation head and all associated relation tails into an instance, allowing the dependencies between span pairs to be modeled. The Entity Subtype classification generates a new input sentence for each extracted entity, where typed markers identify the target entity. This entity-specific version of the sentence feeds into BERT, and the CLS token hidden state feeds into a multi-label classifier consisting of separate linear layers for each \textit{span-with-value} argument. 


\subsection{Evaluation}

Model hyperparameters were tuned using the CAMIR training and validation sets, and the final performance is reported for the withheld CAMIR test set. Performance is presented using the overlap span equivalence criterion, where two spans are considered equivalent if they overlap. For instance, when extracting anatomy spans in line 2 of Figure \ref{annotation_example}, the anatomy span, ``right pedicle'' would be considered equivalent to ``T12 right pedicle" since there is an overlap between the spans. Triggers are considered equivalent if the event types are identical and spans overlap. \textit{Span-only} arguments are considered equivalent if the argument types match, argument spans overlap, and connected triggers are equivalent. \textit{Span-with-value} argument equivalence is similar to the equivalence of \textit{span-only} arguments, except that the subtype labels must also match. Overlap span equivalence is relevant to the CAMIR annotation schema and extraction task as most arguments are normalized to predefined concepts. This overlap criterion is also suited for downstream secondary-use applications, and we performed extensive error analyses to validate this criterion (see Section \ref{strictVSoverlap}). Performance is evaluated using precision, recall, and F1, and statistical significance is calculated using a non-parametric (bootstrap) test \cite{bergkirkpatrick2012}.


\section{Results}

\subsection{Corpus}

This section summarizes CAMIR, including the IAA and distribution of annotations. Table \ref{annotator_agreement} presents IAA for the doubly annotated reports. The overall IAA for all the triggers and arguments in the doubly-annotated reports was 0.762 F1 using the criteria defined in Section 3.3. Consensus regarding the trigger annotation was higher at 0.856, 0.805, and 0.854 F1 for \textit{Indication}, \textit{Lesion}, and \textit{Medical Problem} triggers, respectively. \textit{Size}, \textit{Size Trend}, and \textit{Count} occur much less frequently than the other arguments, contributing to the lower IAA for these arguments. \textit{Characteristic} spans are very linguistically diverse, resulting in frequent false negatives. The double annotation and adjudication of the validation and test sets mitigates the impact of this lower IAA on the evaluation.

\begin{table}[htb!]
    \small
    \centering
    \renewcommand{\arraystretch}{0.9}

\begin{tabular}{llc}\toprule
\textbf{Event type} &\textbf{Argument type} &\textbf{F1} \\\midrule
\multirow{4}{*}{Indication} &Trigger &0.856 \\
&Assertion &0.820 \\
&Anatomy &0.797 \\
&Indication Type &0.804 \\
\midrule

\multirow{7}{*}{Lesion} &Trigger &0.805 \\
&Assertion &0.762 \\
&Anatomy &0.710 \\
&Size &0.715 \\
&Size Trend &0.560 \\
&Count &0.564 \\
&Characteristic &0.481 \\
\midrule
\multirow{3}{*}{Medical Problem} &Trigger &0.854 \\
&Assertion &0.815 \\
&Anatomy &0.751 \\
\midrule
\multirow{1}{*}{Overall} & & 0.762\\
\bottomrule
\end{tabular}

    \caption{Inter-annotator agreement (IAA) for doubly annotated radiology reports (n=357)}
    \label{annotator_agreement}
\end{table}

\begin{table*}[htb!]
    \small
    \centering
    \renewcommand{\arraystretch}{0.9}

\begin{tabular}{lllcccc}\toprule
\multirow{2}{*}{\textbf{Event Type}} &\multirow{2}{*}{\textbf{Argument Type}} &\multirow{2}{*}{\textbf{Argument Subtype}}
&\multicolumn{3}{c}{\textbf{Frequency (avg. per report)}} \\\cmidrule{4-6}
& & &\textbf{CT} &\textbf{MR} &\textbf{PET-CT}

\\\midrule
\multirow{6}{*}{Indication} &Trigger &- &507 (2.5) &496 (2.4) &491 (2.4) \\
\cmidrule{2-6}
&\multirow{3}{*}{Assertion} &present &449 &435 &436 \\
& &absent &11 &1 &5  \\
& &possible &47 &60 &50 \\
\cmidrule{2-6}
&\multirow{4}{*}{Type} &neoplastic dx  &184 &181 &193  \\
& &non-neoplastic dx &112 &102 &91 \\
& &symptom &149 &150 &134 \\
& &trauma &23 &32 &21\\
\cmidrule{2-6}
&Anatomy &all &276 &263 &278  \\
\midrule
\multirow{14}{*}{Lesion} &Trigger &- &1855 (9.2) &1967 (9.7) &1887 (9.3) \\
\cmidrule{2-6}
&\multirow{3}{*}{Assertion} &present &1190 &1302 &1222  \\
& &absent &547 &531 &539  \\
& &possible &118 &134 &126  \\
\cmidrule{2-6}
&Anatomy &all &2321 &2536 &2378 \\
\cmidrule{2-6}
&\multirow{2}{*}{Size} &current&303 &364 &349  \\
& &past &46 &63 &36 \\
\cmidrule{2-6}
&\multirow{5}{*}{Size Trend} &decreasing &26 &38 &36 \\
& &disappear &22 &18 &26  \\
& &increasing &35 &61 &32 \\
& &new &64 &58 &46 \\
& &no change &109 &142 &130 \\
\cmidrule{2-6}
&Count &- &119 &112 &132  \\
\cmidrule{2-6}
&Characteristic &-  &762 &841 &921\\ 
\midrule
\multirow{5}{*}{\makecell[l]{Medical\\ Problem}} &Trigger &- &2063 (10.2) &2111 (10.4) &2080 (10.2)\\
\cmidrule{2-6}
&\multirow{3}{*}{Assertion} &present &1217 &1294 &1189  \\
& &absent &607 &592 &631 \\
& &possible &239 &225 &260  \\
\cmidrule{2-6}
&Anatomy &all &2197 &2316 &2083 \\
\midrule
& & \textbf{Total number of reports (N)}  & 203 & 202 & 204 \\
\bottomrule
\end{tabular}

    \caption{Distribution of the annotated event types and arguments in CAMIR by modality. Numbers in parentheses indicate the average number of triggers per report.}
    \label{annotation_stats}
\end{table*}

Table \ref{annotation_stats} summarizes the distribution of the annotated phenomena in CAMIR. While the focus of the imaging modality may differ, the distribution of annotations is similar across modalities for most argument types. In each report, 2.4-2.5 \textit{Indication} triggers were identified and the reason for the imaging test was mostly \textit{neoplastic diagnosis}, which refers to the abnormal growth of certain tumors. The number of \textit{Lesion} and \textit{Medical Problem} was similar in all three modality types, where most triggers for both events were assigned \textit{Assertion} value \textit{present}. Approximately 9.2-9.7 \textit{Lesion} and 10.2-10.4 \textit{Medical Problem} events were identified on average in each radiology report.

Lesion-specific attributes such as \textit{Characteristic}, \textit{Size}, \textit{Size Trend}, \textit{Count} add supplementary clinical information that might be crucial for interpreting the result of the imaging tests. In addition, we provide \textit{Assertion} values to each event to clearly indicate the absence, possibility, or presence of each finding. 
These \textit{Assertion} labels are very important for creating accurate and comprehensive representations and are relevant to wide range of secondary use cases.
The granularity of CAMIR also provides the opportunity for more advanced multi-modal research by combining text and relevant images.

\subsection{Information Extraction}

\begin{table*}[htb!]
    \small
    \centering
    \renewcommand{\arraystretch}{0.9}

\begin{tabular}{lrrcccccc}\toprule
\multirow{2}{*}{\textbf{Event}} &\multirow{2}{*}{\textbf{Argument}} &\multirow{2}{*}{\textbf{Count}} &\multicolumn{3}{c}{\textbf{mSpERT}} &\multicolumn{3}{c}{\textbf{PL-Marker++}} \\\cmidrule{4-9}
& & &\textbf{P} &\textbf{R} &\textbf{F1} &\textbf{P} &\textbf{R} &\textbf{F1} \\\midrule
\multirow{5}{*}{Indication} &Trigger &285 &0.818 &0.758 &\textbf{0.787} &0.878 &0.705 &0.782 \\
&Assertion &285 &0.816 &0.730 &\textbf{0.770} &0.852 &0.684 &0.759 \\
&Anatomy Parent &157 &0.696 &0.554 &0.617 &0.711 &0.580 &\textbf{0.639} \\
&Anatomy Child &157 &0.675 &0.529 &0.593 &0.711 &0.580 &\textbf{0.639} \\
& Type &262 &0.783 &0.687 &\textbf{0.732} &0.782 &0.683 &0.729 \\
\hline
\multirow{8}{*}{Lesion} &Trigger &1169 &0.859 &0.846 &0.853 &0.880 &0.888 &\textbf{0.884}\textsuperscript{\textdagger} \\
&Assertion &1169 &0.840 &0.810 &0.825 &0.863 &0.870 &\textbf{0.866}\textsuperscript{\textdagger} \\
&Anatomy Parent &1448 &0.753 &0.620 &0.680 &0.769 &0.673 &\textbf{0.718}\textsuperscript{\textdagger} \\
&Anatomy Child &1448 &0.720 &0.586 &0.646 &0.733 &0.642 &\textbf{0.684}\textsuperscript{\textdagger} \\
&Characteristic &652 &0.654 &0.420 &0.512 &0.776 &0.477 &\textbf{0.591}\textsuperscript{\textdagger} \\
&Count &75 &0.833 &0.800 &\textbf{0.816} &0.902 &0.733 &0.809 \\
&Size &294 &0.761 &0.670 &0.713 &0.890 &0.691 &\textbf{0.778}\textsuperscript{\textdagger} \\
&Size Trend &206 &0.720 &0.587 &0.647 &0.795 &0.714 &\textbf{0.752}\textsuperscript{\textdagger} \\
\hline
\multirow{4}{*}{Medical Problem} &Trigger &1271 &0.897 &0.832 &0.863 &0.886 &0.866 &\textbf{0.875} \\
&Assertion &1271 &0.878 &0.802 &0.839 &0.854 &0.834 &\textbf{0.844} \\
&Anatomy Parent &1349 &0.792 &0.623 &\textbf{0.697} &0.752 &0.633 &0.688 \\
&Anatomy Child &1349 &0.725 &0.563 &\textbf{0.633} &0.687 &0.578 &0.628 \\
\midrule
\multicolumn{2}{c}{OVERALL} &12847 &0.798 &0.684 &0.736 &0.805 &0.718 &\textbf{0.759}\textsuperscript{\textdagger} \\
\bottomrule
\end{tabular}

    \caption{Event extraction performance for mSpERT and PL-Marker++ evaluated using overlap criteria on the held-out test set. Higher F1-scores are bolded. \textsuperscript{\textdagger} indicates statistical significance (p $<$ 0.05)}
    \label{results_stats}
\end{table*}

Table 5 summarizes the extraction performance on the held-out CAMIR test set. PL-Marker++ achieved significantly higher overall performance than mSpERT (0.759 F1 vs 0.736 F1). While the performance of mSpERT and PL-Marker++ models were similar for extracting \textit{Indication} and \textit{Medical Problem} triggers and arguments, PL-Marker++ performed significantly better in extracting \textit{Lesion} triggers and all but one argument type. The PL-Marker++ model achieved gains of $+\Delta$0.05 F1 in extracting \textit{Characteristic}, \textit{Size}, and \textit{Size Trend} arguments for \textit{Lesion} events. The overall improved performance of PL-Marker++ can be attributed to the infusion of the trigger and argument location information through all layers of the BERT model.


\section{Discussion}


\subsection{Annotation Quality}

The IAA for CAMIR exceeds 0.70 F1 for most arguments. Exceptions are \textit{Size Trend}, \textit{Count}, and \textit{Characteristic}. We observed that \textit{Size Trend} and \textit{Count} are relatively infrequent in our data set and are therefore easy to overlook during annotation. \textit{Characteristic} was introduced as an inclusive catchall category, resulting in diverse lexical phrasing and semantics, consequently yielding a comparatively low IAA. 

The IAA for \textit{Lesion} and \textit{Medical Problem} triggers was above 0.80 F1. Majority of the remaining disagreements resulted from ambiguity between event types. For example, generic words such as ``disease" can refer to a \textit{Lesion} trigger in the context of ``residual disease," indicating a small number of cancer cells. At the same time, ``disease" can refer to a \textit{Medical Problem} in the context of ``small vessel disease".  Similarly, ``recurrence" can be either \textit{Lesion} or \textit{Medical Problem} depending on the finding that is recurring.

\subsection{Model Performance}

Table \ref{results_stats} shows the BERT models achieved the highest performance for the extraction of triggers and some of the more regularly-expressed arguments such as \textit{Count}. 
\textit{Anatomy} is a crucial argument for capturing the meaning of the radiology reports and has an extraction performance of 0.628-0.718 F1, indicating further study is needed to improve extraction performance.

\subsection{Strict vs Overlap Evaluation}

\label{strictVSoverlap}
To validate the span overlap criterion, we evaluated the performance of PL-Marker++ on event triggers using a strict, exact match span criterion. This evaluation resulted in  test set performance of 0.749 F1 for \textit{Indication}, 0.681 F1 for \textit{Lesion}, and 0.765 F1 for \textit{Medical Problem} triggers. There were 279 triggers that were equivalent using the overlap criterion but not equivalent using exact match. We manually reviewed these trigger predictions to assess their clinical meaning relative to the reference triggers. For all 279 of these discrepancies between the overlap and strict criterion, the predicted triggers still captured all information important to identifying clinical findings. We found that 203 of these trigger predictions were shorter than the reference, often omitting modifiers (e.g. reference - “Mild FDG activity” vs. predicted - “FDG activity” or reference - “hypodense lesions” vs. predicted - “lesions”) and 76 trigger predictions were longer than the reference, often including modifiers (e.g. reference - “lesion” vs. predicted “mass lesion” or reference “carcinoma” vs. predicted “renal cell carcinoma”). 

\subsection{Generalizability of the Annotation and Extraction Performance}
Our annotation guidelines are designed to be comprehensive and foundational to derive overall clinical findings from radiology reports. The guidelines do not rely on specific templates or formats used in our institution. Even though the structure of the medical imaging reports may differ across modalities or institutions, we expect the description of the main clinical findings in the reports to be compatible with our annotation guidelines. Moreover, although our annotation focused on three imaging modalities, the annotation schema is not specific to particular modalities. Therefore, we anticipate that minimal modifications will be required to the annotation schema to create annotated datasets at different institutions or for other modality types, if any. 
However, since the content and linguistics may vary among institutions and modality types, directly using information extraction models trained on CAMIR may achieve lower performance on the reports at other institutions or for other modalities. Domain adaptation of the CAMIR-trained models may be required, to maintain high performance.

\section{Conclusion}

\renewcommand{\thefootnote}{\fnsymbol{footnote}}


We introduce a novel annotated corpus, CAMIR, consisting of CT, MRI, and PET-CT reports from a large hospital system. CAMIR has been annotated using a granular event schema, where clinical indication, lesion, and medical problem findings are captured through multiple arguments and most arguments are normalized to predefined radiological concepts. Using CAMIR, we explored two BERT-based architectures (1) mSpERT, an existing system which jointly extracts all event information, and (2) PL-Marker++, a system that extracts the event information through multiple stages, which we augmented before applying to CAMIR. These systems performed comparable to IAA. Our PL-Marker++ achieved significantly higher overall performance than mSpERT (0.759 F1 vs 0.736 F1).  These systems show that the fine-grained information in CAMIR can be reliably extracted by automatic methods. While these systems perform well overall, triggers and their assertion arguments are more reliably extracted than other arguments such as anatomy.  The annotation guidelines for CAMIR and the source code for the IE models presented in this paper are available on our GitHub repository\footnote{https://github.com/uw-bionlp/CAMIR}. CAMIR is unique in that it combines clinical concept normalization with the granularity of relation/event annotations to produce comprehensive semantic representations that can easily be incorporated into secondary-use applications, including clinical decision support \cite{demner2009can}, surveillance \cite{haas2005use}, follow-up tracking \cite{mabotuwana2019automated}, report simplification \cite{qenam2017text}, cross-specialty diagnosis correlation \cite{filice2019deep}, and automated impression generation \cite{wiggins2021natural}. 

\section{Limitations}
This study is limited to data from a single urban hospital system and focuses on three imaging modalities. While CAMIR includes more than 13,000 clinical events, it only consists of 609 reports. Therefore, the generalizability of the annotated corpus and extraction architectures to other hospital systems and other imaging modalities needs further exploration. In future work, we will incorporate additional modalities such as radiographs, ultrasound, and mammography. Additionally, we will evaluate the performance of larger generative Large Language Models (e.g. GPT4) in fine-tuning and in-context learning settings.

\section{Acknowledgements}

This work was supported by the National Institutes of Health (NIH) - National Cancer Institute (Grant Nr. 1R01CA248422-01A1) and National Library of Medicine (NLM) Biomedical and Health Informatics Training Program at the University of Washington (Grant Nr. T15LM007442). The content is solely the responsibility of the authors and does not necessarily represent the official views of the NIH. 

\section{Ethics}

We obtained the necessary approvals from our institution's IRB, with a waiver of patient consent to use their clinical notes. Radiology reports may contain patient Protected Health Information (PHI), like names, contact information, and other identifiers. Each report was automatically de-identified using a neural de-identification model and then subsequently manually de-identified by medical student annotators, to ensure no remaining PHI. All radiology reports, including the original and de-identified versions, were stored on a Health Insurance Portability and Accountability Act (HIPAA)-compliant server, to ensure patient privacy. All researchers and annotators received the necessary human subjects training to interact with patient data, including PHI.

 The annotated reports in our corpus were randomly sampled from the general population of patients with medical imaging from a single institution. The demographics of the patients were not considered during data collection, and the patient populations in our corpus may not be representative of populations at other institutions or the broader population, which may inadvertently bias the distribution of annotated medical conditions.  Additionally, radiology reports of other institutions may differ in format and language. These factors may impact the generalizability of the extraction models developed using the corpus. 
 



\bibliographystyle{lrec-coling2024-natbib.bst}
\bibliography{lrec-coling2024-example.bib}


\onecolumn

\section*{Appendix A. SNOMED-CT Concepts for Anatomy Normalization}

\begin{table}[ht]
    \renewcommand{\thetable}{A1}
    \renewcommand{\arraystretch}{0.5}
    \vspace{0.05in}
    \centering
    {\small 
    
\begin{tabular}{m{3.5cm} m{9.0cm} m{1.2cm}}

\toprule
\textbf{Anatomy Parent} &
  \textbf{Anatomy Children} &
  \textbf{Count} 
   \\
\toprule
\raggedright{Abdomen (113345001)} &
  Abdominal Wall (83908009), Adrenal Gland (23451007), Mesentery (89679009), Peritoneal Sac (118762006), Retroperitoneal (699600004), Spleen (78961009), Undetermined &
  512 
   \\
   \midrule
\raggedright{Cardiovascular System (59820001)} &
  Arterial (51114001), Coronary Artery (41801008), Heart (80891009), Pericardial Sac (76848001), Pulmonary Artery (81040000), Venous (119553000), Undetermined &
  770 
   \\
   \midrule
\raggedright{Digestive System (49596003)} &
  Esophagus (32849002), Intestine (113276009), Large Intestine (14742008), Small Intestine (30315005), Stomach (69695003), Undetermined &
  425 
   \\
   \midrule
\raggedright{Female Reproductive System (27436002) \& Obstetric (308762002)} &
  Adnexal (23043003), Breast (76752008), Extra-embryonic (314908006), Female Genital Structure (53065001), Fetus (55460000), Ovary (15497006), Placenta (78067005), Umbilical Cord (29870000), Uterus (35039007), Undetermined &
  272 
   \\
   \midrule
\raggedright{Head \& Neck (774007)} &
  Ear (117590005), Eye (371398005), Laryngeal (4596009), Mouth (385294005), Nasal Sinus (2095001), Neck (45048000), Pharynx (54066008), Thyroid (69748006), Undetermined &
  1096 
   \\
   \midrule
\raggedright{Hepato-Biliary System (34707002, 122489005)} &
  Bile Duct (28273000), Gallbladder (28231008), Liver (10200004), Pancreas (15776009), Undetermined &
  609 
   \\
   \midrule
\raggedright{Lymphatic (91688001)} &
  Undetermined &
  559 
   \\
   \midrule
\raggedright{Male Reproductive System (90264002)} &
  Epididymis (87644002), Prostate (119231001), Testis (40689003), Undetermined &
  49 
   \\
   \midrule
\raggedright{Miscellaneous} &
  Adipose Tissue (55603005), Biomedical Device (63653004), Connective Tissue (21793004), Undetermined &
  59 
   \\
   \midrule
\raggedright{Musculoskeletal (312717002)} &
  Bone/Joint, Skeletal and or Muscle (71616004), Undetermined &
  1811 
   \\
   \midrule
\raggedright{Neurological System (25087005)} &
  Brain (12738006), Cerebrospinal Fluid Pathway (280371009), Cerebrovascular System (28661005), Extraaxial (1231004), Nerve (3057000), Pituitary (56329008), Spine Cervical (122494005), Spine Cord (2748008), Spine Lumbar (122496007), Spine Sacral (699698002), Spine Thoracic (122495006), Spine Unspecified (421060004), Undetermined &
  3235 
   \\
   \midrule
\raggedright{Other Body Regions (272625005)} &
  Entire Body (38266002), Lower Limb (61685007), Pelvis (12921003), Upper Limb (53120007), Undetermined &
  887 
   \\
   \midrule
\raggedright{Respiratory System (714323000)} &
  Lung (39607008), Pleural Membrane (3120008), Tracheobronchial (91724006), Undetermined &
  1200 
   \\
   \midrule
\raggedright{Skin (400199006)} &
  Skin and or Mucous Membrane (707861009), Subcutaneous (71966008), Undetermined &
  58 
   \\
   \midrule
\raggedright{Thoracic (51185008)} &
  Mediastinum (72410000), Undetermined &
  772 
   \\
   \midrule
\raggedright{Urinary System (122489005)} &
  Kidney (64033007), Ureter (119220009), Urinary Bladder (89837001), Undetermined &
  378 \\
\bottomrule
\end{tabular}
    }

    \caption{Anatomy Parent-Child SNOMED Hierarchy. SNOMED concept names are shortened due to lack of space. There are 16 Parent and 71 Child labels. Undetermined Child labels are catch-all categories. Count represents the number of annotations for Parent labels.}
    \label{SNOMED_Concepts}
\end{table}


\end{document}